\documentclass[11pt,a4paper]{article}
\usepackage{lipsum}
\usepackage{authblk}
\usepackage[top=1in, bottom=1in, left=1in, right=1in]{geometry}
\usepackage{fancyhdr}
\usepackage{multicol}
\usepackage{graphicx}
\usepackage{amssymb}
\usepackage{amsmath}
\usepackage{url}
\usepackage{color}
\usepackage{times}
\usepackage{multirow}

\renewenvironment{abstract}{%
\hfill\begin{minipage}{0.95\textwidth}
\rule{\textwidth}{1pt}}
{\par\noindent\rule{\textwidth}{1pt}\end{minipage}}

\makeatletter
\renewcommand\@maketitle{%
\hfill
\begin{minipage}{0.95\textwidth}
\vskip 2em
\let\footnote\thanks 
{\LARGE \@title \par }
\vskip 1.5em
{\large \@author \par}
\end{minipage}
\vskip 1em \par
}
\makeatother
\begin{document}

\fancypagestyle{plain}{%
  \renewcommand{\headrulewidth}{0pt}%
  \fancyhf{}%
  \fancyhead[C]{The final version of this paper has been published in Neurocomputing journal, Elsevier; available via http://www.sciencedirect.com. Please cite this paper as: \\``\textcolor{red}{Amirhossein Tavanaei and Anthony Maida, A Spiking Network that Learns to Extract Spike Signatures from Speech Signals, \textit{Neurocomputing}, 240, 191-199, 2017}"}%
}
%
\title{\textbf{A Spiking Network that Learns to Extract Spike Signatures from Speech Signals}}
\author[]{Amirhossein Tavanaei}
\author[]{Anthony S. Maida}
\affil[]{The Center for Advanced Computer Studies \\
University of Louisiana at Lafayette, LA 70504, USA \\
tavanaei@louisiana.edu, maida@cacs.louisiana.edu}
\maketitle
\begin{abstract}
\textbf{Abstract} \\
Spiking neural networks (SNNs) with adaptive synapses reflect core properties of biological neural networks. 
Speech recognition, as an application involving audio coding and dynamic learning, provides a good test problem
to study SNN functionality. 
We present a simple, novel,
and efficient 
nonrecurrent SNN that learns to convert a speech signal into a spike train signature.
The signature is distinguishable from signatures for other speech signals representing different words,
thereby enabling digit recognition and discrimination in devices that use only spiking neurons.
The method uses a small, nonrecurrent SNN consisting of Izhikevich neurons equipped with spike timing dependent plasticity (STDP)
and biologically realistic synapses.
This approach introduces an efficient and fast network without error-feedback training, although it does require supervised training. 
The new simulation results produce discriminative spike train patterns for spoken digits in which highly correlated spike trains 
belong to the same category and low correlated patterns belong to  different categories. 
The proposed SNN is evaluated using a 
spoken digit recognition task where a subset of the Aurora speech dataset is used. 
The experimental results show that the network performs well in terms of accuracy rate and complexity.
\\


\end{abstract}
\\

\begin{multicols}{2}
\section{Introduction}

Spiking neural networks (SNNs) with adaptive synapses reflect core properties of nearly all biological 
networks.
An important mechanism of Hebbian synaptic modification in biological networks is known as
spike-timing-dependent plasticity (STDP) \cite{Caporale2008a,markram2011}.
STDP-type
mechanisms take into account the relative spike times of pre- and postsynaptic neural spikes
to adjust the strength of a synapse connecting two neurons.
The question of what STDP accomplishes in a learning framework is, and has been, under intense investigation.
Spiking neurons and STDP learning rules have been applied in diverse fields of pattern 
recognition and classification \cite{Kasabov2013a,Kasinski2006a,Storck2001a,Wysoski2010a,Panchev2004a,Wang2014a} 
such as learning and information processing of visual 
features \cite{Masquelier2007a,tavanaei2016b,Wysoski2008a}  
and speech recognition \cite{Wade2010a,tavanaei2016}. 

Our work studies the performance of a novel STDP-trained,
nonrecurrent SNN for isolated spoken digit recognition.
The spike trains produced by the output neurons in this network have discriminative properties.
That is,
the spike signature of an output neuron contains substantial information about the digit presented to the network.
The net input to these neurons, which drives their output spikes, can be used
to train a support-vector machine (SVM) to recognize the presented digit.
The information encoded in a spike train is an example of temporal coding.

The learning is applied to the output neurons and
uses a mixture of Hebbian and anti-Hebbian STDP in a supervised fashion.
Specifically, if an output unit is being trained to recognize the spoken digit ``one,''
then it undergoes Hebbian STDP when an exemplar of ``one'' is presented and anti-Hebbian STDP otherwise.


Our proposed architecture is
a small feedforward network of spiking neurons 
that is trained by
using the combination of supervised Hebbian and anti-Hebbian STDP just described.
The learning has two effects.
First, the net inputs to the output neurons can be used to train an SVM for spoken digit recognition.
Second, the output spike trains have discriminative properties and, in principle, could
be used to perform the classification task.


The small network 
is efficient and can be trained (or used) quickly, while showing promising accuracy.  
Also, the trained synaptic weights extract input signatures invariant to different speakers (male and female) 
and signal variants.

\subsection{Related work}
We discuss two approaches to using spiking neural networks in spoken digit classification.
The first approach uses feedforward-architectures because our network is a single-layer feedforward network.
The second approach uses a recurrent
architecture in the form of a liquid state machine (LSM)\@.

\subsubsection{Comparable work}

Much research has studied neurocomputational approaches to ASR mimicking the biological inspiration of the human auditory 
system \cite{Dibazar2004a,Nager2002a,Namarvar2001a}.
The auditory system has components for encoding the raw signal (inner ear) and generating appropriate spike trains 
(cochlea).

Schafer and Jin~\cite{schafer2014} developed a 
template-based, single-layer network architecture for spoken digit recognition.
Its emphasis is on recognizing noise-corrupted, spoken digits.
The network consists of 32 input units fully connected to up to 1,100 output units.
The input units are driven by a bank of 32 cochlear gammatone filters that process the 
speech signal.
Each
output unit is separately trained by a support-vector machine (SVM) to respond to a
particular preprocessed acoustic feature and, in response to speech input,
generates a spike train.
The collection of spike trains from each of the output neurons compose a spike raster, which is taken as the network output.
The output spike raster is compared to prototype rasters using a longest-common-substring (LCS)
algorithm to classify the input signal.
To support different pronunciations and signal variations, they used up to 100 prototype-templates per digit. 
They reported 82\% to 99\% accuracy rates for the networks using a range of 1 to 100 templates per digit. 
Our work differs from theirs in two important ways.
First,
our work uses
a biologically plausible STDP algorithm to train the output units.
Second, we use ten output units, in contrast to 1,100, and each of our units
detects one digit.
Their output units detect speech formant features and our output units detect digits.
We use an output signature of a single neuron, instead of a spike raster of features,
to recognize a digit.


Wade et al~\cite{Wade2010a} introduced a learning method that merges the STDP rule 
with the BCM \cite{bienenstock1982} learning rule (so that the acquired weights are more stable).
Their spiking network was two-layer with 5040 neurons in the hidden layer and ten neurons in the output layer,
representing each of the ten digit classes.
In the output layer, the neuron with the highest firing rate determines the classification decision.
This contrasts with our network where the output neuron's spike signature is more relevant to the
classification decision.
They also used a highly speculative
global weight mapping rule (hypothesized to be mediated by astrocytes)
to control the relative occurrence of similar data patterns across classes. 
They evaluated their model using different benchmark problems including spoken digit recognition. 
The SNN proposed by Wade et al. consists of frequency-selective filters followed by a layer which undergoes local learning. 
This network uses 50,400 adaptive synapses, so it is a much larger network than ours (which uses 2,000 synapses). 

Dibazar et al.\
proposed a 
feature extraction method using a continuous dynamic synaptic neural network 
to implement a biologically plausible network
for spoken digit recognition by a classifier~\cite{Dibazar2003a,Dibazar2004a}.
They achieved 99\% and 40\% accuracy rates for clean and noisy (10 dB) signals respectively, but 
at the expense of high computational complexity. Later, they developed a biologically plausible discrete dynamic NN to extract features from the speech signal with 85\% and 45\% accuracy rates for clean and noisy (10 dB) spoken digits. 
In this architecture, much of the information is encoded in the real-time dynamics of the synapses.
Our network uses static, adjustable synapses so the network operation is fundamentally different.

Dao et al~\cite{dao2014} introduced a sparsity-based representation for spoken digit recognition. 
Although they did
not use SNNs, the goal for fast processing and effective signal discrimination for pattern recognition was accomplished.

\subsubsection{Reservoir-based approaches}
Several studies have used reservoir-based approaches to perform spoken digit classification 
\cite{Verstraeten2005a,Verstraeten2006a,Schrauwen2007a,Harris2007a,Ghani2008a}.
See \cite{Jaeger2009a} for a review of reservoir-based approaches in general.
As a whole,
this work has been quite successful in achieving near perfect digit recognition performance
with robustness to noise.
Reservoirs were proposed as a solution to the slow convergence of recurrent neural networks, which were
of interest because of their ability to store temporal information.
Reservoir computing avoids the convergence issue by using a suitably structured RNN as a reservoir (temporal memory)
which is not trained.
A good reservoir operate on the edge chaos and implements a fading memory.
The reservoir can consist of either spiking \cite{Verstraeten2005a} or non-spiking \cite{Verstraeten2006a} neurons.
Training is reserved for a linear readout layer that trains rapidly.
The linear readout performs well because the RNN maps the inputs to a higher dimensional space in which
the categories are more likely to be linearly separable.

Although the reservoir may or may not be built from spiking neurons, to our knowledge,
in the context of speech recognition,
there is only one study that use a
trainable readout layer with spiking neurons \cite{zhang2015}.
In most studies, the state of a spiking reservoir is low-pass filtered \cite{Verstraeten2005a,Verstraeten2006a}
before being sent to the readout layer.
This allows the readout layer to be rapidly trained using any traditional non-spiking method for a single-layer architecture \cite{Ghani2008a}.
Zhang et al~\cite{zhang2015} is the only study that
presented an LSM in which the readout layer was trained using a bio-inspired, spike-based learning rule. 

Although the reservoir approach described above yields excellent performance on spoken digit recognition,
the question of training a spiking network in the context of speech recognition is not addressed 
in this research. 
Also, the operation of reservoir is computationally more expensive than a single-layer, feed-forward SNN\@.
The present paper explores the training of a single layer SNN for extracting the spike train signatures 
from the spoken digits.
Similar to an LSM, our network does train a classifier by
using the net inputs to the readout neurons.



\section{Feature Extraction}
\label{sec:featureExtraction}
Feature extraction converts a raw signal into a more usable form. 
The speech signal is divided into small overlapping time sections called speech frames. 
The Hamming window, which is commonly used in discrete time signal processing, is used in signal framing due to its frequency features~\cite{Oppenheim}.
Our SNN needs a fixed number of frames, $N$, for each spoken digit. As the length, $L$, of a spoken digit can vary from 500 to 1000 ms, we divided it into $N=40$ frames with 50\% overlap to 
support a frame length of 10-50 ms. 
The 50\% overlap ($\gamma = 0.5$) captures the temporal characteristics of the changing spectrum of the speech signal.
The window size (frame length) in milliseconds is calculated based on $L$, $N$, and $\gamma$, as shown below.

\begin{equation}
\mathit{window\ size_{(\mathrm{ms})}}=\frac{L _{(\mathrm{ms})}}{N(1-\gamma)+\gamma}
\label{eqn:windowSize}
\end{equation}

\noindent
Because window size increases with signal duration, spoken words pronounced slowly have longer frames in comparison to words pronounced quickly. 

After framing, a small feature vector for each frame is extracted.
There are several methods for speech feature extraction such as MFCC and Mel-scaled discrete wavelet coefficient (MFDWC)~\cite{Tavanaei2011a}. 
We instead use a minimal feature vector extracted from the frame's frequency spectrum, as explained below.

\subsection{Frequency spectrum}

\begin{figure*}
\center

\includegraphics[viewport=0.6in 5.2in 8.4in 8.3in,clip=true,scale=.6]{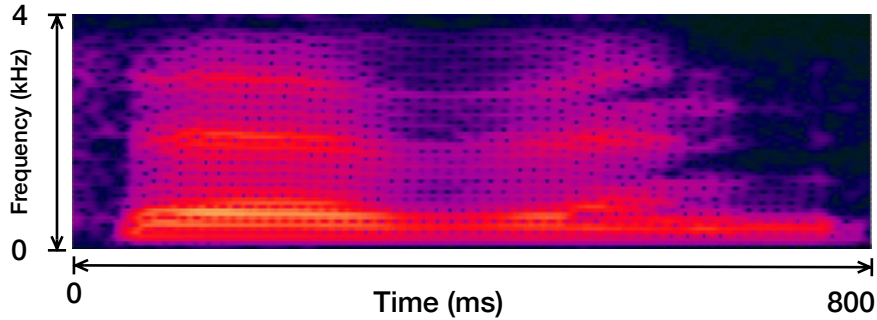}

\caption{Spectrogram for the digit $seven$. The horizontal axis shows time (500-1000 ms word length) and vertical axis shows the frequency 
which increases from bottom (0 Hz) to top (4000 Hz).
Color shows energy with yellow $>$ red $>$ blue. The first frequency component (DC value) is in the range 10 to 50 Hz.}
\label{fig:spectrogram}
\end{figure*}

As a speech signal unfolds in time, the power of its frequency spectrum varies.
This can be visualized in a spectrogram as shown in Fig.~\ref{fig:spectrogram}.
Spectrograms can be used to identify spoken words phonetically, and to analyze the audio files in specific frames. 
Spectrum calculation of a frame is shown in Eq.~(\ref{eqn:spectrogram}). The spectrum values are calculated for all the frames temporally to represent the speech signal spectrogram.
Fig.~\ref{fig:spectrogram} shows the spectrogram for the spoken digit $seven$.
\begin{equation}
\mathit{Spectrum}=\log{\mathit{|FFT(frame)|}^2}
\label{eqn:spectrogram}
\end{equation}
\subsection{Frequency band}
Low frequencies in the spectrogram have more energy and information relevant to classification than the high frequencies (cf. Fig.~\ref{fig:spectrogram}). 
Thus, an effective feature vector provides more resolution in low frequencies. This is obtained by using incrementally spaced frequency bands. 
We create the frequency bands from the Fibonacci sequence. 
This sequence provides good frequency band sizes for a small number of features
(other approaches to doing this are also viable).

\begin{figure*}
\center
\includegraphics[width=12cm]{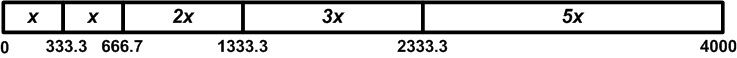}
\caption{Five frequency bands for 0-4000 Hz frequency range.}
\label{fig:frequencyBands}
\end{figure*}

A separate feature vector is calculated for each frame. 
A frame encompassing an $R$ Hz frequency range can be divided into $M$ frequency bands. 
In this paper, $R=4000$ Hz and $M=5$. 
The number of filter banks ($M=5$) is small enough to create a minimal SNN. Although more feature values characterize the speech frame with higher resolution, the large input vector increases the network's computations. 
Therefore, five filter banks are sufficient for this purpose. 
Additionally, for the isolated spoken digit recognition problem, three filter banks extracting the acoustic features in the range 0 to 1500 Hz are able to represent only eight vowels and one nasal phoneme pronounced in the spoken digits (`aa' as one, `u' as two, `ee' as three and zero, `o' as four and zero, `ai' as five and nine, `i' as six, `e' as seven, `ei' as eight, and n).   
If the first frequency band length is $x$, then the filter bank containing $M=5$ bands will 
have lengths of $x, x, 2x,3x, 5x$. 
Specifically, each frame represented in the $R=4000$ Hz frequency range is divided into $M=5$ bands with lengths 
of (333.3, 333.3, 666.7, 1000, and 1666.7) as shown in Fig.~\ref{fig:frequencyBands}.
$x$ is chosen so that the equality below is satisfied.
\begin{equation}
R = \sum_{i=1}^5 \mathrm{fib}(i) \cdot x = 12 x
\end{equation}

The value of each element in a feature vector is the average energy over the range given
in Fig.~\ref{fig:frequencyBands} for a given frame.
For example,
the first feature value codes the average energy in the range $0 - 333.3$ Hz.

\section{Input Spike Generation}
\label{sec:inputSpikeGeneration}
\begin{figure*}
\center

\includegraphics[viewport=0.5in 0.0in 12.5in 8.0in, clip=true, scale=.21]{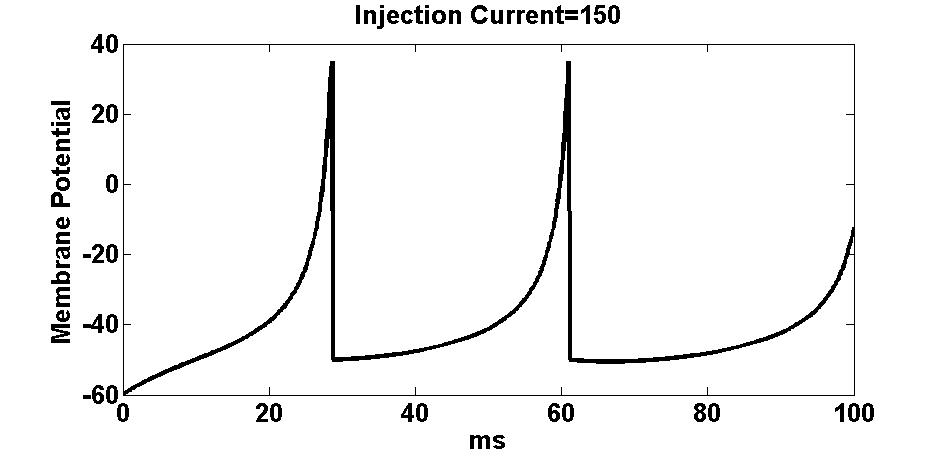} 
\quad
\includegraphics[viewport=0.5in 0.0in 12.5in 8.0in, clip=true, scale=.21]{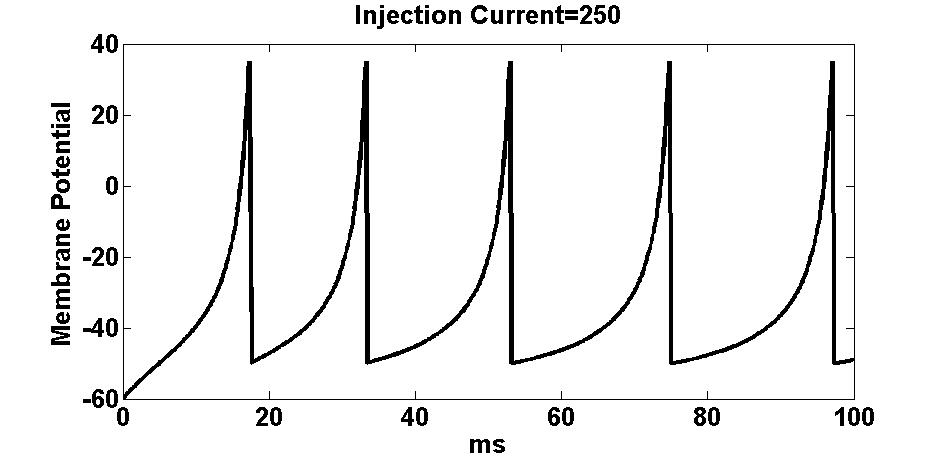} 

\caption{Spike coding: RS neuron spikes with $I_\mathrm{inj}$ equal to 150 (left) and 250 (right) for a duration of $T=100$ ms.
Both plots show spike-rate adaptation to constant input.}
\label{fig:spikes}
\end{figure*}

We use the Izhikevich model regular spiking (RS) neuron  \cite{Izhikevich2003a}
to convert a feature component to a spike train, 
as seen in Eqs.~(\ref{eqn:Izhikevich1}) through~(\ref{eqn:Izhikevich3}).
Extracted features control the value of the injected input current, $I_{\mathrm{inj}}$, to an afferent $y$ unit.
The $I_{\mathrm{inj}}$ drives the system. For the $y$ units, the only input is $I_\mathrm{inj}$, so $I_\mathrm{inj}=I_\mathrm{tot}$ in the equations below.
A larger total current causes more frequent spikes as seen in Fig.~\ref{fig:spikes}.
Also, it can be seen that the neurons exhibit spike-rate adaptation to a constant input
(which is a common characteristic of biological neurons).
\begin{equation}
C\frac{dV}{dt}=k(V-V_{\mathrm{rest}})(V-V_{\mathrm{th}})-U+I_{\mathrm{tot}}
\label{eqn:Izhikevich1}
\end{equation}
\begin{equation}
\frac{dU}{dt}=a[b(V-V_{\mathrm{rest}})-U]
\label{eqn:Izhikevich2}
\end{equation}
and the reset equation
\begin{equation}
\mathrm{if}\ V>V_{\mathrm{peak}}:V=c,\ U=U+d,\ \mathrm{Spike\ is\ emitted}
\label{eqn:Izhikevich3}
\end{equation}

\noindent
The spike time is the time step at which the membrane potential, $V$, becomes greater than $V_\mathrm{peak}$.
$U$ specifies a recovery factor inhibiting the spike and keeps the membrane potential near
the resting value, $V_{\mathrm{rest}}$. 
The neuron capacity ($C$), threshold ($V_{\mathrm{th}}$), $V_{\mathrm{peak}}$, and symbols $a,$ $b,$ $k,$ $c,$ $d$ are constants,
specified in Table~\ref{tab:RSparams},  
whose values 
control the dynamic characteristics of the system and cause the neuron to have regular spiking behavior. 

Each feature vector component drives one RS neuron by controlling the value of $I_\mathrm{inj}$
($y$ unit, as explained in the next section), causing it to generate a spike train over a fixed 
duration $T=100$ milliseconds.

\section{Network Architecture}
\begin{figure*}
\centering

\includegraphics[viewport=1.6in 6.2in 8.4in 9.3in,clip=true,scale=1]{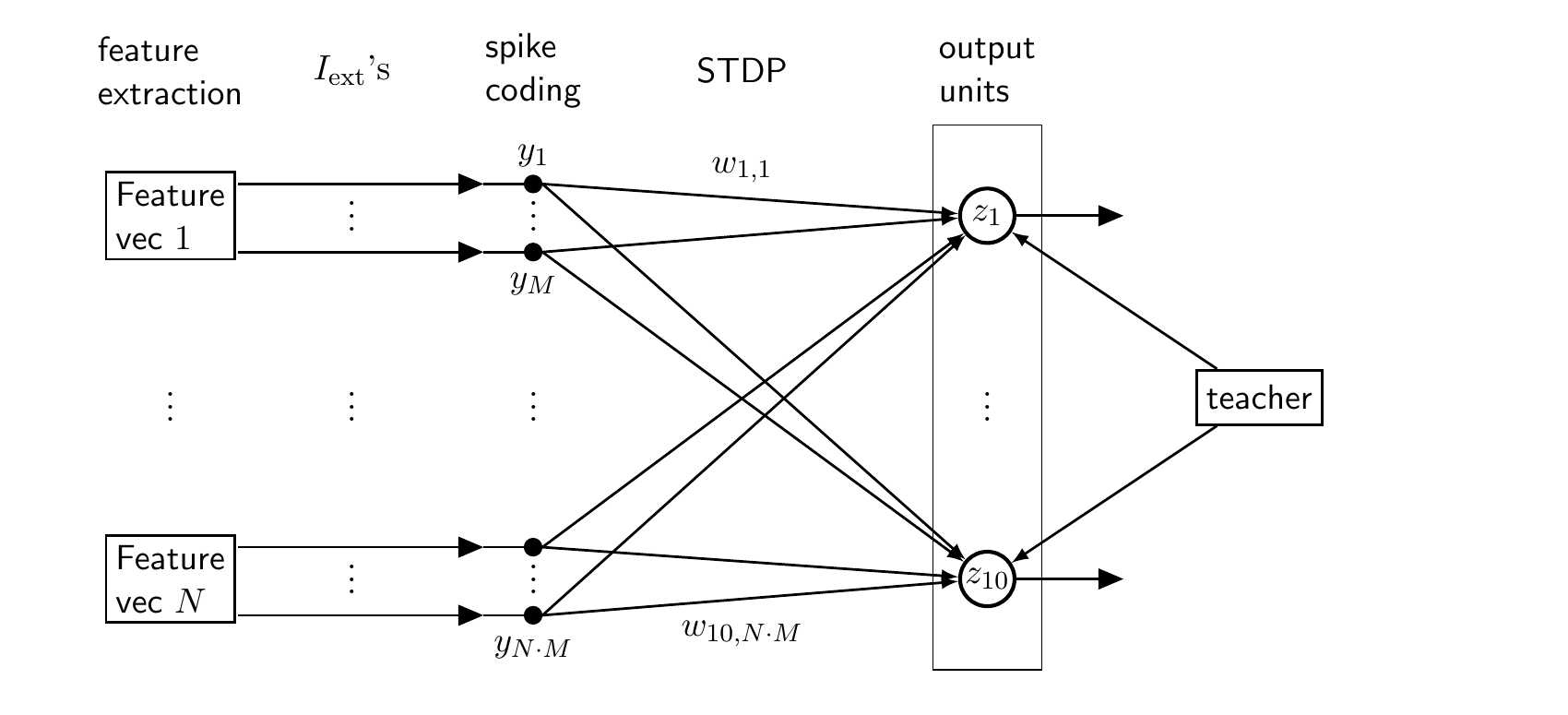}

\caption{Network architecture. 
Each of the $N$ feature vectors produced by a feature extraction box has $M=5$ components.
Each component becomes $I_{inj}$ to a $y$ unit which then converts it to a spike train.
The number of y units equals $N\cdot M = 200$.
Output units $z_1 - z_{10}$ represent the ten digit categories.
The teacher signal controls whether Hebbian STDP or anti-Hebbian STDP is applied.
$N\cdot M$ synapses project to each $z$ unit. 
There are $10\cdot N\cdot M = 2000$ trainable synapses.}
\label{fig:network}
\end{figure*}

Our network is trained using STDP with labeled data.
After training, the network can generate spike train signatures for the ten digit categories.
The test signatures can be compared with spike trains from target data to perform classification.
The network architecture appears in Fig.~\ref{fig:network}. 
It consists of: 
\begin{enumerate}
\item For training,
      the network input consists of feature vector input from $N=40$ frames.
      The $N$ frames cover the duration of the speech input stream.
      Each feature vector has $M=5$ components as described in Sec.~\ref{sec:featureExtraction}.
      For training, the sequential input is buffered and then presented to the network 
      simultaneously (For testing, the procedure is slightly different).
      At first glance, it might seem like this might cause the sequential dependencies in the input to be lost.
      However, this is not the case. 
      The sequence information is simply converted from a temporal to a geometric format.      
      Although the input is pooled for training, the raw sequential information is preserved when 
      generating spike signatures after training (explained in Sec.~\ref{signature}).
\item The feature values are given to $y$ units that are implemented as RS neurons (configured according 
      to Table~\ref{tab:RSparams}).
      Each $y$ unit accepts one of five feature vector components which serves as
      its $I_\mathrm{inj}=I_\mathrm{tot}$ input value as described in Sec.~\ref{sec:inputSpikeGeneration}.
      There are a total of $N\cdot M = 200$ $y$ units.
\item An output layer of ten $z$ units is used.
      Each unit corresponds to one of the ten spoken digit categories (class labels).
      These are also implemented as RS neurons with the same parameter configuation as the $y$ units (Table~\ref{tab:RSparams}).
      Their input consists entirely of synaptic input from the 200 $y$ units, namely $I_\mathrm{syn}=I_\mathrm{tot}$.
      The $y$ units are fully connected to the $z$ units.
      The $z$ units are trained according to the procedure described in Sec.~\ref{sec:learning}.
\item Finally, there is a teacher that monitors the $z$ units in order to determine the form of the STDP\@ used in training.
      If the target unit spikes at a given time step, it undergoes case 1 of Hebbian STDP and the rest of the (nontarget) units undergo case 1 of anti-Hebbian STDP\@.
      If the desired unit does not spike, it undergoes case 2 of Hebbian STDP and the rest of the units undergo case 2 of anti-Hebbian STDP\@.
	  The teaching signal is only used for the training phase.
\end{enumerate}

\begin{table*}
\footnotesize
\caption{RS neuron parameters for both $y$ and $z$ units.}
\label{tab:RSparams}
\centering
\begin{tabular}{|l|l|l|l|}
\hline
Parameter  & Value        & Parameter & Value             \\ \hline
$V_{\mathrm{rest}}$ & \textbf{-60} & $a$         & \textbf{0.03}     \\ 
$V_{\mathrm{th}}$   & \textbf{-40} & $b$         & \textbf{-2}       \\ 
$V_{\mathrm{peak}}$ & \textbf{35}  & $c$         & \textbf{-50}      \\ 
$C$         & \textbf{100} & $d$         & \textbf{100}      \\ 
$K$          & \textbf{0.7} & $U_0$      & \textbf{0}        \\ 
$\Delta T$   & \textbf{0.1} & $I_{\mathrm{inj}}$ & \textit{variable} \\ \hline
\end{tabular}
\end{table*}



\section{Learning}
\label{sec:learning}
There are two types of learning in the model.
The most important type is the STDP that is used to train 
the synapses projecting to the output units.
The other type of learning occurs after the synapses are trained.
The net input to the trained output units is used to train an SVM for classification
(explained in Sec. \ref{subsec:ClassPerfResults}).

\subsection{Neuron Model}
\begin{figure*}
\center
\includegraphics[width=8cm]{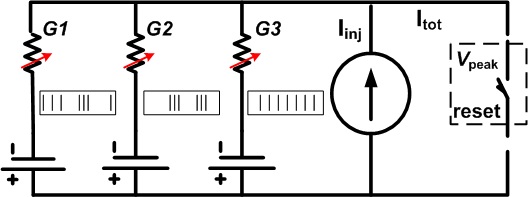}\includegraphics[width=6cm]{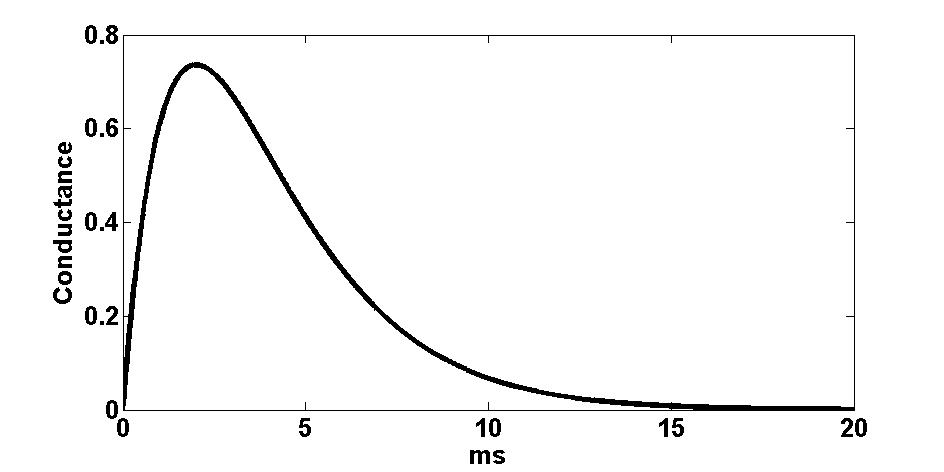}
\caption{Left: Computation of $I_{\mathrm{syn}}$ with three synaptic inputs, showing $I_{\mathrm{syn}}$, $I_{\mathrm{inj}}$, and $I_{\mathrm{tot}}$. 
Right: Graph of synaptic conductance change over time after receiving a spike, $K_{\mathrm{syn}}$=1 and $\tau$=2.}
\label{fig:circuit}
\end{figure*}

Fig.~\ref{fig:circuit} (left) shows the simulation circuit of a model neuron.
The dashed box represents the spike generation step described in 
Eqs.~(\ref{eqn:Izhikevich1} -- \ref{eqn:Izhikevich3}).
The branches marked $G_1$ to $G_3$ represent synaptic conductances for three synapses.
If the neuron is a $y$ unit, then there is no synaptic input, only
injected current $I_\mathrm{inj}$.
In this case, $I_\mathrm{tot}=I_\mathrm{inj}$.
For a $z$ unit, there is synaptic input $I_\mathrm{syn}$, but no injected current.
In this case, $I_\mathrm{tot}=I_\mathrm{syn}$.
Each $z$ unit has $N\cdot M=200$ incoming synapses, corresponding to the 200 afferent $y$ units.

Learning occurs by modifying the synaptic conductances.
$G_k(t)$ 
denotes the synaptic conductance change over time for synapse $k$ caused by receiving a single input spike to that synapse. 
The $\alpha$--function (Eq.~\ref{eqn:alpaFun}) models the conductance time-course of the synapse. 
Fig.~\ref{fig:circuit} (right) shows the $\alpha$--function graph for one synapse receiving one spike at time $t$.
\begin{equation}
G(t) = K_{\mathrm{syn}} \cdot t \cdot e^{-t/\tau}
\label{eqn:alpaFun}
\end{equation}
$K_{\mathrm{syn}}$ controls the conductance amplitude.
This is what is adjusted during learning.
Synaptic weight adjustments change the value of $K_\mathrm{syn}$
according to Eq.~(\ref{eqn:deltaKsyn}).
$\tau$ is the time at which the synapse reaches its maximum conductance. 
$t$ represents the elapsed time since the most recently received spike.

When multiple spikes are received in succession before a conductance drops to zero, the successive conductance effects are added linearly
according to Eq.~(\ref{eqn:Gtot}).
Specifically, the total conductance of $N\cdot M$ input synapses with $N_{\mathrm{rec},k}$ ($k=1:N\cdot M$) spikes is calculated by
summing linearly over the synapses and input spikes:

\begin{equation}
G_{\mathrm{tot}}=\sum_{k=1}^{N\cdot M}\sum_{j=1}^{N_{\mathrm{rec},k}}K_{\mathrm{syn},k}(t-t_{k,j}^\mathrm{f})e^{-(t-t_{k,j}^\mathrm{f})/\tau}
\label{eqn:Gtot}
\end{equation}

\noindent
where $t^\mathrm{f}_{k,j}$ is the spike time of spike $j$ for synapse $k$.
$N_{\mathrm{rec},k}$ denotes the number of spikes received by synapse $k$.
The total synaptic current $I_\mathrm{tot}$ is given by:
\begin{equation}
I_{\mathrm{syn}}(t)=\sum_{k=1}^{N\cdot M}E_{\mathrm{syn},k}G_{\mathrm{syn},k}^{\mathrm{tot}}(t)-V(t)\sum_{k=1}^{N\cdot M}G_{\mathrm{syn},k}^{\mathrm{tot}}(t)
\label{eqn:Isyn}
\end{equation}

\noindent
In our simulations, $E_{\mathrm{syn},k}=0$.

\subsection{Spike Timing Dependent Plasticity (STDP)}
\label{subsec:STDP}
Weight adjustment at a synapse is governed by the relative spike times of its pre- and 
postsynaptic neurons (Eq.~\ref{eqn:mainSTDP}) in conjunction with the teacher feedback. 
The teacher feedback dictates the form of the STDP, whether it be Hebbian or anti-Hebbian.
In the case of normal Hebbian STDP,
if the postsynaptic spike is generated immediately after receiving the presynaptic spike, 
the presynaptic spike has a causal role in the output neuron firing. 
The synaptic weight is thus increased (LTP). 
Conversely, if a postsynaptic spike occurs before the presynaptic spike, the strength is reduced (LTD), as seen in the equation below.
\begin{equation}
\Delta w_{ji}=
\begin{cases}
0.01Ae^{\frac{-(|t_j^f-t_i^f|)}{\tau+}} & t_j^f-t_i^f\geq 0 \ , \ \ A>0\\
0.01Be^{\frac{-(|t_j^f-t_i^f|)}{\tau-}} & t_j^f-t_i^f<0 \ , \ \ B<0
\end{cases}
\label{eqn:mainSTDP} 
\end{equation}

\noindent
In the above, the first case (Case 1) covers LTP and the second case (Case 2) covers LTD\@.
Both cases are decaying exponentials that decay with the distance between and pre- and postsynaptic spikes.
$A > 0$ and $B < 0$ scale the amplitude of the exponential, and 
$\tau+$ and $\tau-$ are the respective time constants. 

Eq.~(\ref{eqn:mainSTDP}) describes Hebbian STDP\@.
To obtain anti-Hebbian STDP, we swap the cases.
The teacher determines which $z$ units undergo Hebbian versus anti-Hebbian STDP\@.
During training, whenever a $z$ unit emits a spike, it undergoes some form of STDP\@.
If the $z$ unit represents the target category, then it undergoes Hebbian STDP\@.
Otherwise, it undergoes anti-Hebbian STDP\@.


The synaptic weight change contributes to a change in the conductance amplitude, $K_\mathrm{syn}$, in the $\alpha$-function model. 
We link the weight adjustment to the adjustment of $K_{ji}$, used in Eq.~\ref{eqn:alpaFun},
by using the equation below.

\begin{equation}
\Delta K_{ji}=\Delta w_{ji}K_{ji}
\label{eqn:deltaKsyn} 
\end{equation}

We now summarize the simulation's operation during training.
The simulation is advanced using $\Delta t = 0.1$ ms time steps using forward Euler (which is adequate for this problem).
The $y$ and $z$ units are updated in a manner consistent with a feedforward sweep.
Whenever a $z$ unit fires, the teacher determines which variant of STDP to apply for that unit.
We also renormalize the weights, using $L_1$, after each training sample.

%
%
\subsection{Obtaining spike signatures}
\label{signature}
Spike signatures are obtained after training.
To obtain a spike signature from an input sample, 
each input frame is processed individually and sequentially, rather than simultaneously (as was done with training).
A frame, 
which contains 5 feature values, is converted to a spike train (with $T=5$ ms and $\Delta T=0.1$ ms) 
by passing through the corresponding inputs to the trained network. 
Each of the forty frames is
passed through the network sequentially such that a spike signature of an idealized duration equal to 200 ms is obtained (40$\times$5 ms). 

\section{Experimental Methods and Results}
\subsection{Data Preparation}
Our experiments were conducted on the Aurora dataset of isolated
spoken digits recorded from different male and female
speakers \cite{Pearce2000a}.
The dataset was used for three purposes: training the network (500 samples), testing the trained
network without noise (500 samples), and testing the trained network with noise (500 noisy samples, SNR=10~dB).

For training,
500 spoken digit samples, with 50 representatives for each digit (0 -- 9), 
were randomly sampled from the dataset.
Each sample was divided into 40 frames with 50\%
overlap Hamming windows, according to Eq. (\ref{eqn:windowSize}).
The feature vector for a frame was obtained by applying the Fourier transform to the wave data
and calculating the average energy of the five Fibonacci-scaled bands.
This produced $N=40$ feature vectors of $M=5$ components.
These were concatenated into a global feature vector of $N\cdot M=200$ components.
The feature vector values form the $I_\mathrm{inj}$ input to the $y$ units shown in Fig.~\ref{fig:network}.

\subsection{Training}
\label{subsub:training}
Before training the weights were initialized to uniform random values between 0 and
1 and then normalized using the $L_1$ norm.
For each training sample, the network operated as follows.
The global feature vector for that sample was presented to the $y$ units for a duration of $T=100$ milliseconds.
The $y$ units generated spikes from their respective $I_\mathrm{inj}$ input as shown in Fig.~\ref{fig:spikes}.
Each of the ten output neurons received 200 spike trains of duration 100 ms via 200 trainable synapses.
The 2,000 incoming synapses to the $z$ unit layer were trained such that 200 synapses representing the presented
digit category underwent Hebbian STDP and the remaining synapses underwent anti-Hebbian STDP\@.

\begin{figure*}
\center
\includegraphics[width=13cm]{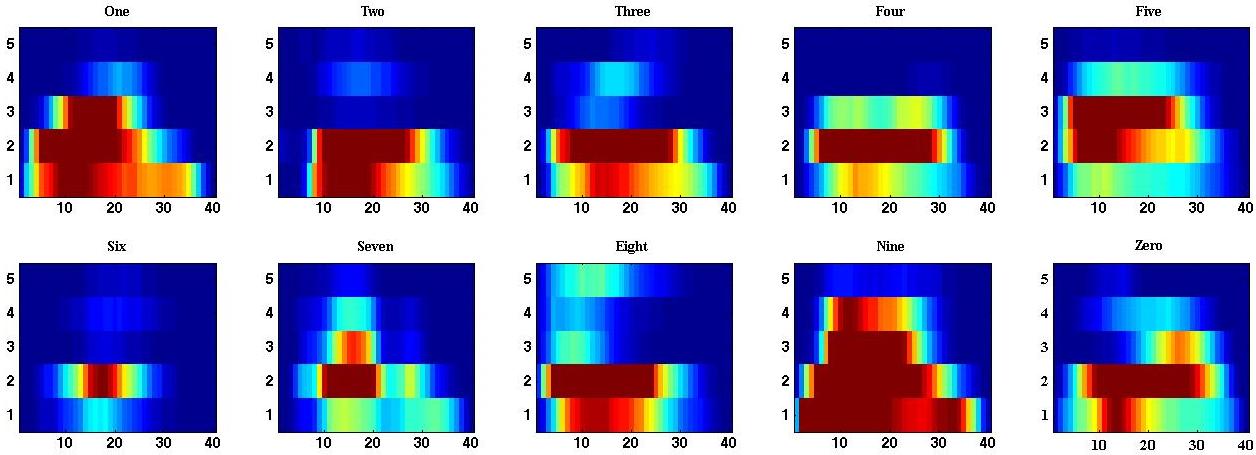}
\caption{Trained synaptic weights 
connecting all $y$ units to each of the ten $z$ units. Each image is 40 (number of frames) 
by 5 (number of features in each frame, low frequencies at bottom) which represents the 200 incoming 
synaptic weights of a particular $z$ unit. (yellow $>$ red $>$ blue)}
\label{fig:synweights}
\end{figure*}

The synaptic weights were renormalized after the presentation of each training example. 
Because convergence was rapid,
training was stopped after 100 epochs, each of which consisted of 500 training samples. 
Fig.~\ref{fig:synweights} shows the trained synaptic weights arranged so that they can be compared with a spectrogram
like that shown in Fig.~\ref{fig:spectrogram}. 
Each point $(f,v)$ in this figure shows the synapse passing a spike train with respect to frame $f$ (in the range 1 to 40) and 
feature value $v$ (in the range 1 to 5). 

\subsection{Testing method}
\label{sec:testingMethod}
In testing mode the network generates spike signatures.
Both \textit{prototype} and \textit{test} signatures are generated.
Test spike signatures are produced by an output unit by submitting a spoken digit sample
to the network after it has been trained.
Prototype spike signatures represent the response of an output to an `average' exemplar for 
the digit class.

\subsubsection{Prototype spike signatures}
We first explain how prototype spike signatures are generated.
These signatures are generated for each of the ten digit categories after training.
The trained network is used to generate these signatures.
Before generating a prototype spike signature for a given category, we create
a representative input feature set for that category.
This involves reusing the training data. 
Specifically,
we average over the extracted feature coefficients for each digit class
in the training set (50 samples per class)
to obtain representative feature input for that class.
That yields $40\times5$ values for each of ten new representative samples.
These ten new items are given to the trained network to generate a prototype spike signature
for each category (according to section~\ref{signature}).
As explained before,
each of the forty frames is
passed through the network sequentially such that a spike signature of an idealized duration equal to 200 ms is obtained (40$\times$5 ms). 
The prototype spike trains appear in Figs.~\ref{fig:targetAndTestSpikeTrainsBeforeTraining} (left) and \ref{fig:targetAndTestSpikeTrains} (left) before and after training respectively.


\subsection{Spike signature results}

\begin{figure*}
\center
\includegraphics[viewport=0.8in 0.0in 9.0in 10.0in,clip=true,scale=0.33]{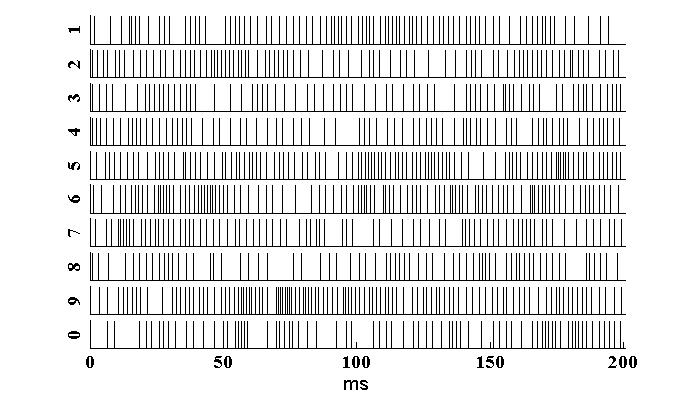}
\quad
\includegraphics[viewport=0.8in 0.0in 9.0in 8.0in,clip=true,scale=0.33]{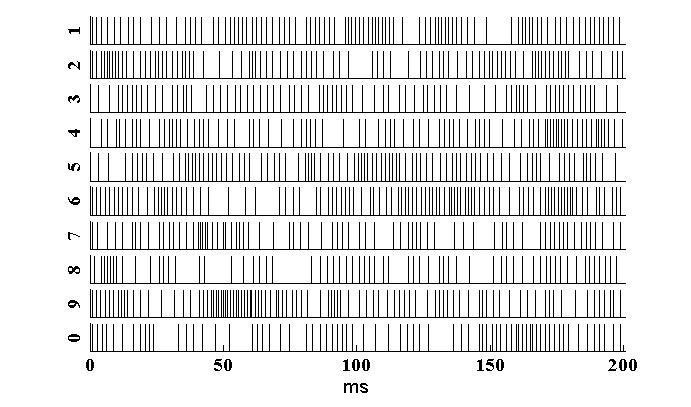}
\caption{
Before training.
Prototype spike trains based on 50 samples per class (left) and 
example test spike trains (right) for randomly selected spoken digits 0 to 9.
Each spike train has a duration of $T=200$ ms.}
\label{fig:targetAndTestSpikeTrainsBeforeTraining}
\end{figure*}

\begin{figure*}
\center
\includegraphics[viewport=0.0in 0.0in 5.0in 5.0in,clip=true,scale=0.53]{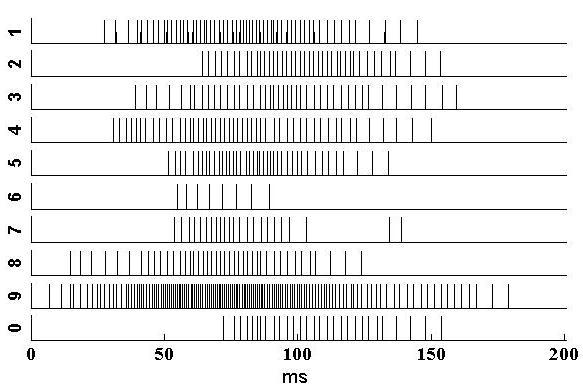}
\quad
\includegraphics[viewport=0.0in 0.0in 5.0in 5.0in,clip=true,scale=0.53]{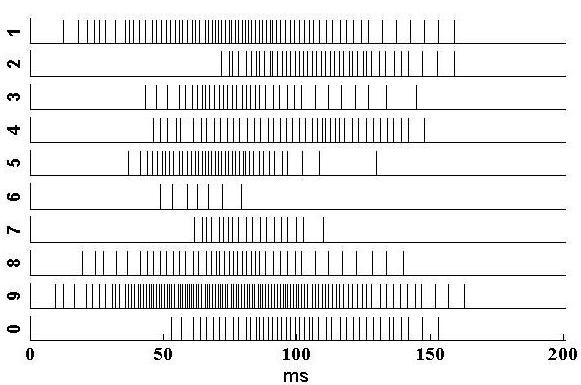}
\caption{After training.
Protype spike trains based on 50 samples per class (left) and 
example test spike trains (right) for randomly selected spoken digits 0 to 9.
Each spike train has a duration of $T=200$ ms.}
\label{fig:targetAndTestSpikeTrains}
\end{figure*}

A set of spoken digits 0 to 9, not used in training,
was randomly selected to obtain test spike signatures. 
Testing spike signatures were generated analogously to prototype spike signatures, 
however, using a single test sample as input for eqch test signature. 
The resulting test spike trains after training appear in Fig.~\ref{fig:targetAndTestSpikeTrains} (right).
Each test signature corresponds to a single randomly selected spoken digit.
Fig.~\ref{fig:targetAndTestSpikeTrainsBeforeTraining} shows spike train signatures before network training.
The spikes are roughly uniformly distributed and dense.
Comparison with Fig.~\ref{fig:targetAndTestSpikeTrains} after training shows that meaningful spike signatures emerge.
Comparison between spike signatures in Fig.~\ref{fig:targetAndTestSpikeTrains} (left and right)
shows that signatures for randomly selected test digits
resemble the corresponding prototype signatures. 
Emitted spikes for the same digits show similar temporal patterns visually. 
For example, the test signature for a digit six in Fig.~\ref{fig:targetAndTestSpikeTrains} (left) is similar to 
the
prototype for category six (right). 
Specifically, its temporal patterns are similar where they have uniformly distributed spikes between 40 to 80 ms. 
However, the temporal patterns of the other digits (0-5, 7-9) have a large distance from the digit six target signature.
To quantify this, 
we use the Victor–-Purpura distance metric \cite{victor1997,victor2005} that quantifies dissimilarity of spike trains. 
Table~\ref{tab:victor} shows the distances between target and prototype signatures calculated by 
the spike interval 
metric\footnote{\url{www-users.med.cornell.edu/~jdvicto/pubalgor.html}} 
\cite{victor1997}. 
The spoken digits 3, 4, and 0 have not been distinguished as accurately as the other digits.   

\begin{table*}[]
\centering
\footnotesize
\caption{Distance calculated for the target and test signatures using the Victor–-Purpura metric. 
Each column shows dissimilarity values between the prototype signature and a randomly selected test signature
for each category.}
\vspace{12pt}
\label{tab:victor}
\begin{tabular}{|l|l|l|l|l|l|l|l|l|l|l|}
\hline
\multicolumn{1}{|c|}{\multirow{2}{*}{\textbf{Prototype}}} & \multicolumn{10}{c|}{\textbf{Randomly selected test}}                                                                                       \\ \cline{2-11} 
\multicolumn{1}{|c|}{}                                  & \textbf{1} & \textbf{2} & \textbf{3} & \textbf{4} & \textbf{5} & \textbf{6} & \textbf{7} & \textbf{8} & \textbf{9} & \textbf{0} \\ \hline
\textbf{1}                                              & 27         & 33         & 42         & 52         & 26         & 57         & 50         & 37         & 73         & 36         \\ \hline
\textbf{2}                                              & 40         & 17         & 29         & 35         & 30         & 41         & 37         & 27         & 80         & 22         \\ \hline
\textbf{3}                                              & 51         & 33         & 26         & 27         & 44         & 41         & 30         & 24         & 96         & 23         \\ \hline
\textbf{4}                                              & 43         & 23         & 31         & 33         & 38         & 48         & 36         & 29         & 90         & 22         \\ \hline
\textbf{5}                                              & 33         & 27         & 33         & 43         & 20         & 43         & 33         & 34         & 77         & 33         \\ \hline
\textbf{6}                                              & 65         & 43         & 30         & 43         & 42         & 6          & 21         & 38         & 100        & 41         \\ \hline
\textbf{7}                                              & 54         & 28         & 22         & 27         & 32         & 26         & 14         & 29         & 97         & 27         \\ \hline
\textbf{8}                                              & 44         & 27         & 26         & 35         & 30         & 41         & 34         & 21         & 82         & 22         \\ \hline
\textbf{9}                                              & 110        & 115        & 123        & 129        & 106        & 137        & 125        & 125        & 67         & 115        \\ \hline
\textbf{0}                                              & 51         & 31         & 22         & 30         & 36         & 30         & 20         & 29         & 99         & 28         \\ \hline
\end{tabular}
\end{table*}

\subsection{Classification performance results}
\label{subsec:ClassPerfResults}
A natural approach to classifying spoken digits would be to match its test signature against
the set of class prototypes and choose the class with the closest match.
Unfortunately, the performance of this approach was not that good.
That is, the prototypes obtained from the class average of the input features was not
sufficiently precise to support good classification.
Instead, a different classification method was implemented that used the net input to
the output neurons.

For classification, the net input (obtained from a single exemplar)
to an output unit was used to train an SVM\@.
The net input to an output unit was subdivided into forty, 5-millisecond duration frames
corresponding to the 200 millisecond input signal.
The net input was assumed to be a good indicator of the number of neural spikes
generated by that output unit (possibly with a small time delay).
In summary,
feature vectors used for training the classifier preserved temporal information at the resolution
of 40 bins and 5 ms per bin.
The net input corresponds to $I_\mathrm{tot}$ in Fig.~\ref{fig:circuit} in which five
input synapses are used (the number of features in a frame).


The performance of the classifier is shown in Table~\ref{tab:performancySummary}.
Overall accuracy is 91 percent for the clean test stimuli.
Accuracy drops to 70 percent when noisy stimuli are used. 
To provide more information, specifically, on what types of errors were made,
Tables~\ref{tabl:confuseNonoise2} and \ref{tabl:confuseWithNoise2} show confusion matrices 
for the clean and noisy conditions, respectively.

\begin{table*}[]
\centering
\footnotesize
\caption{Overall performance accuracy.}
\vspace{12pt}
\label{tab:performancySummary}
\begin{tabular}{|l|l|l|}
\hline
\textbf{Measure}                    & \textbf{No Noise} & \textbf{10 dB Noise} \\ \hline
Average Hit Ratio (\%)              & 90.9              & 70.9                 \\ \hline
Average Misclassification Rate (\%) & 9.3               & 29.9                 \\ \hline
Overall Accuracy                    & 90.8              & 70.2                 \\ \hline
\end{tabular}
\end{table*}

\begin{table*}[]
\centering
\scriptsize
\caption{Confusion matrix obtained using net input method for spoken digit recognition without noise. 
500 unused samples were used for this test.
Overall accuracy was 91\%, calculated by summing along the diagonal to count the number of correct answers and dividing
by 500.
Off-diagonal rows entries indicate number of misses for that target.
Off-diagonal columns entries indicate number of detection errors.
Bottom right entry is diagonal total.}
\label{tabl:confuseNonoise2}
\vspace{12pt}
\begin{tabular}{|l|r|r|r|r|r|r|r|r|r|r|r|r|}
\hline
\multicolumn{1}{|l|}{\textbf{Desired}}&\multicolumn{10}{|c|}{\textbf{Recognized}}&\multicolumn{1}{l|}{Row}&\multicolumn{1}{l|}{Hit}\\\cline{2-11}
\textbf{digit}  & \multicolumn{1}{c|}{\textit{\textbf{1}}} & \multicolumn{1}{c|}{\textit{\textbf{2}}} & \multicolumn{1}{c|}{\textit{\textbf{3}}} & \multicolumn{1}{c|}{\textit{\textbf{4}}} & \multicolumn{1}{c|}{\textit{\textbf{5}}} & \multicolumn{1}{c|}{\textit{\textbf{6}}} & \multicolumn{1}{c|}{\textit{\textbf{7}}} & \multicolumn{1}{c|}{\textit{\textbf{8}}} & \multicolumn{1}{c|}{\textit{\textbf{9}}} & \multicolumn{1}{c|}{\textit{\textbf{0}}} & \multicolumn{1}{l|}{totals} & \multicolumn{1}{l|}{rate (\%)} \\ \hline
\textit{\textbf{1}} & 45   & 0   & 0    & 0   & 0   & 0    & 1    & 0   & 3    & 0   & 49  & 91.8 \\ \hline
\textit{\textbf{2}} & 0    & 50  & 3    & 0   & 0   & 2    & 0    & 2   & 0    & 1   & 58  & 86.2 \\ \hline
\textit{\textbf{3}} & 1    & 2   & 45   & 1   & 0   & 1    & 1    & 0   & 2    & 0   & 53  & 84.9 \\ \hline
\textit{\textbf{4}} & 1    & 0   & 0    & 47  & 0   & 2    & 4    & 0   & 0    & 0   & 54  & 87.0 \\ \hline
\textit{\textbf{5}} & 0    & 0   & 0    & 0   & 48  & 0    & 0    & 0   & 0    & 0   & 48  & 100  \\ \hline
\textit{\textbf{6}} & 0    & 0   & 0    & 0   & 0   & 42   & 0    & 2   & 0    & 1   & 45  & 93.3 \\ \hline
\textit{\textbf{7}} & 1    & 1   & 2    & 1   & 0   & 0    & 39   & 1   & 0    & 0   & 45  & 86.7 \\ \hline
\textit{\textbf{8}} & 0    & 1   & 0    & 0   & 0   & 2    & 0    & 52  & 0    & 0   & 55  & 94.5 \\ \hline
\textit{\textbf{9}} & 1    & 0   & 0    & 0   & 1   & 0    & 0    & 0   & 41   & 2   & 45  & 91.1 \\ \hline
\textit{\textbf{0}} & 1    & 0   & 0    & 2   & 0   & 0    & 0    & 0   & 0    & 45  & 48  & 93.8 \\ \hline
Column totals                  & 50   & 54  & 50   & 51  & 49  & 49   & 45   & 57  & 46   & 49  & 500 &      \\ \hline
Miss rate (\%)                  & 10.0 & 7.4 & 10.0 & 7.8 & 2.0 & 14.3 & 13.3 & 8.8 & 10.9 & 8.2 &     & 454  \\ \hline
\end{tabular}
\end{table*}

\begin{table*}[]
\centering
\scriptsize
\caption{Confusion matrix obtained using net input method for spoken digit recognition with noise SNR=10 dB. 
500 unused noisy samples were used for this test.
Overall accuracy was 70\%, calculated by summing along the diagonal to count the number of correct answers and dividing
by 500.
Off-diagonal row entries indicate number of misses for that target.
Off-diagonal column entries indicate number of detection errors.
Bottom right entry is diagonal total.}
\label{tabl:confuseWithNoise2}
\vspace{12pt}
\begin{tabular}{|l|r|r|r|r|r|r|r|r|r|r|r|r|}
\hline
\multicolumn{1}{|l|}{\textbf{Desired}}&\multicolumn{10}{|c|}{\textbf{Recognized}}&\multicolumn{1}{l|}{Row}&\multicolumn{1}{l|}{Hit}\\\cline{2-11}
\textbf{digit}  & \multicolumn{1}{c|}{\textit{\textbf{1}}} & \multicolumn{1}{c|}{\textit{\textbf{2}}} & \multicolumn{1}{c|}{\textit{\textbf{3}}} & \multicolumn{1}{c|}{\textit{\textbf{4}}} & \multicolumn{1}{c|}{\textit{\textbf{5}}} & \multicolumn{1}{c|}{\textit{\textbf{6}}} & \multicolumn{1}{c|}{\textit{\textbf{7}}} & \multicolumn{1}{c|}{\textit{\textbf{8}}} & \multicolumn{1}{c|}{\textit{\textbf{9}}} & \multicolumn{1}{c|}{\textit{\textbf{0}}} & \multicolumn{1}{l|}{totals} & \multicolumn{1}{l|}{rate (\%)} \\ \hline
\textit{\textbf{1}} & 38   & 0    & 1    & 3    & 2    & 1    & 3    & 0    & 8    & 3    & 59  & 64.4 \\ \hline
\textit{\textbf{2}} & 1    & 35   & 10   & 1    & 0    & 5    & 2    & 2    & 1    & 2    & 59  & 59.3 \\ \hline
\textit{\textbf{3}} & 2    & 6    & 34   & 0    & 0    & 2    & 3    & 5    & 3    & 1    & 56  & 60.7 \\ \hline
\textit{\textbf{4}} & 2    & 1    & 2    & 39   & 0    & 8    & 4    & 3    & 0    & 2    & 61  & 63.9 \\ \hline
\textit{\textbf{5}} & 1    & 0    & 0    & 0    & 45   & 0    & 1    & 0    & 5    & 0    & 52  & 86.5 \\ \hline
\textit{\textbf{6}} & 0    & 3    & 2    & 2    & 0    & 25   & 0    & 2    & 0    & 1    & 35  & 71.4 \\ \hline
\textit{\textbf{7}} & 4    & 3    & 3    & 1    & 0    & 0    & 28   & 0    & 2    & 0    & 41  & 68.3 \\ \hline
\textit{\textbf{8}} & 0    & 2    & 2    & 1    & 0    & 5    & 1    & 35   & 0    & 0    & 46  & 76.1 \\ \hline
\textit{\textbf{9}} & 4    & 0    & 0    & 0    & 2    & 0    & 1    & 0    & 32   & 1    & 40  & 80.0 \\ \hline
\textit{\textbf{0}} & 2    & 1    & 0    & 3    & 1    & 2    & 1    & 1    & 0    & 40   & 51  & 78.4 \\ \hline
Column totals                  & 54   & 51   & 54   & 50   & 50   & 48   & 44   & 48   & 51   & 50   & 500 &      \\ \hline
Miss rate (\%)                 & 29.6 & 31.4 & 37.0 & 22.0 & 10.0 & 47.9 & 36.4 & 27.1 & 37.3 & 20.0 &     & 351  \\ \hline
\end{tabular}
\end{table*}

\subsection{Comparison to other approaches}
Our approach with 91\% accuracy compares favorably with other 
recent investigations regarding spike-based neural networks for spoken digit recognition in terms of 
the combined factors of network complexity and accuracy rate. 
Table~\ref{tab:compare} presents a rough comparison of accuracy achieved in other recent studies. 
The Aurora data set is roughly comparable to the TI46 data set~\cite{doddington1981}.
Aurora is based on a version of the TIDigits data set, but downsampled at 8 kHz  using an `ideal' low-pass filter.
The clean spoken digits are then distorted artificially.


\begin{table*}[]
\centering
\footnotesize
\caption{Performances reported for the spoken digit recognition task using SNN and sparse representation. 
The network in \cite{Dibazar2004a} is trained and evaluated only on the spoken digits zero through three of TIDigits speech corpus.}
\vspace{12pt}
\label{tab:compare}
\begin{tabular}{|l|l|l|l|l|}
\hline
\multicolumn{1}{|c|}{\textbf{Model}}          & \multicolumn{1}{c|}{\textbf{Accuracy (\%)}} & \multicolumn{1}{c|}{\textbf{\# Adaptive weights}} & \multicolumn{1}{c|}{\textbf{Dataset}} &
\multicolumn{1}{|c|}{\textbf{\# Speakers}} \\ \hline
Schaffer \& Jin~\cite{schafer2014}      & 82-99                                       & $32\cdot 1100=35,200$                              & Aurora & 100                               \\ \hline
Dao et al~\cite{dao2014}   & 95                                          & Not SNN &     Aurora    & --                        \\ \hline
Wade et al~\cite{Wade2010a}                  & 95.25                                       & $5040\cdot\ 10 = 50,400$                   & TI46                                 & 16 \\ \hline
Verstraeten et al~\cite{Verstraeten2005a}      & 97.5, 99.5 (Best)                                        & $(300$ to $1900)\cdot 10> 3000$                        & TI46  & 5                                \\ \hline
Zhang et al~\cite{zhang2015}             & 92.3, 99                                        & $135\cdot 10=1350$                      & TI46                                 &16, 5 \\ \hline
Dibazar et al~\cite{Dibazar2004a} & 85.5                                        & $45\cdot 4\cdot 10=1800$                        & TIDigits & --                                \\ \hline
Our Model                                     & 91                                          & $200\cdot 10=2000$                               & Aurora & 50                               \\ \hline
\end{tabular}
\end{table*}  




\section{Discussion and conclusion}
\label{sec:discussion}
Our model represents
a novel method for creating spike train signatures from spoken digits.
It uses the Izhikevich RS neuron model 
combined with STDP learning. 
The learning was implemented by Hebbian and anti-Hebbian STDP controlled by a teaching signal.
The trained network produced 
target spike signatures for the ten spoken digits. 
Prototype signatures obtained from a set of average input feature values
produced similar signatures for the same categories and different signatures for the non-similar categories. 
The proposed SNN provided a fast spike signature extraction system for both male and female speech signals.
A signature-based classifier obtained 91\% and 70\% overall accuracy 
rates in categorizing the clean and noisy spoken digits, respectively. 

Small, spike-based networks, when appropriately adapted, enable power efficient implementations on neuromorphic hardware. 
The proposed single-layer SNN uses a small number of spiking neurons and adaptive synapses to implement a fast and efficient model to extract spike signatures for spoken digits. 
The 
filter banks 
extract only five feature values for each frame to create a minimal network while performing reasonably. 
The Hebbian and anti-Hebbian STDP rules adjust the synaptic weights such that the spatio-temporal features of the speech signal are preserved. 
The spike signatures extracted for the digits represent different spiking patterns for different digits. 
The visual comparisons and the distance measures between the prototype and the test spike signatures showed 
the network has power to discriminate the spoken digits. 
Additionally, the classification results (91\% accuracy) were consistent with the characteristics
of the spike signatures. 
Furthermore, the minimal SNN recognized the noisy spoken digits (10 dB) with 70\% accuracy. 

Biological networks at least in part use temporal spike codes, as exemplified by the
spike signatures we have generated in the present study.
The outputs of our network can be used as inputs for 
further processing such as to identify common digit sequences (e.g., ``911'') or more
general modules for word-phrase processing.

Therefore, the proposed network is a useful architecture of spiking neurons to extract feature sets to be used for classification problems. 
In the future investigations, this network can be assigned as the first layer (component) of a spiking deep neural network when its feature 
maps are used as an input set of spike trains for the next layers of spiking neurons.
A limitation of the proposed method comes from the trade-off with its complexity and performance. 
The small feature vector used in this model does not extract fine details of
the signal and would not be a sufficiently sensitive detector for larger vocabularies.




  \bibliographystyle{unsrt} 
  \bibliography{spikeSignatures}

\end{multicols}
\end{document}